\documentclass[runningheads]{llncs}
\usepackage{graphicx}
\usepackage{url}
\usepackage{cite}
\usepackage{amsmath}

\newcommand*{\textastdbltight}{\raisebox{-0.1ex}{*}\llap{\raisebox{-1.25ex}{*}}}

\begin{document}
\title{Unsupervised correspondence with combined geometric learning and imaging for radiotherapy applications}
\titlerunning{Unsupervised correspondence with combined geometric learning and imaging}
%
\author{Edward G. A. Henderson\inst{1}\orcidID{0000-0003-3752-4054} \and
Marcel van Herk\inst{1,2}\orcidID{0000-0001-6448-898X} \and
Andrew F. Green\inst{3}\orcidID{0000-0002-8297-0953} \and
Eliana M. Vasquez Osorio\inst{1,2}\orcidID{0000-0003-0741-994X}}
\authorrunning{E. G. A. Henderson et al.}
%
\institute{The University of Manchester, Oxford Rd, Manchester M13 9PL, UK \and Radiotherapy Related Research, The Christie NHS Foundation Trust, Manchester M20 4BX, UK \and European Bioinformatics Institute, EMBL-EBI, Cambridge, UK\\
\email{edward.henderson@postgrad.manchester.ac.uk}\\
}
\maketitle              
\begin{abstract}
The aim of this study was to develop a model to accurately identify corresponding points between organ segmentations of different patients for radiotherapy applications. A model for simultaneous correspondence and interpolation estimation in 3D shapes was trained with head and neck organ segmentations from planning CT scans. We then extended the original model to incorporate imaging information using two approaches: 1) extracting features directly from image patches, and 2) including the mean square error between patches as part of the loss function. The correspondence and interpolation performance were evaluated using the geodesic error, chamfer distance and conformal distortion metrics, as well as distances between anatomical landmarks. Each of the models produced significantly better correspondences than the baseline non-rigid registration approach. The original model performed similarly to the model with direct inclusion of image features. The best performing model configuration incorporated imaging information as part of the loss function which produced more anatomically plausible correspondences.  We will use the best performing model to identify corresponding anatomical points on organs to improve spatial normalisation, an important step in outcome modelling, or as an initialisation for anatomically informed registrations. All our code is publicly available at \url{https://github.com/rrr-uom-projects/Unsup-RT-Corr-Net}.
\end{abstract}

\section{Introduction}
Radiotherapy is used in the treatment of $\sim80\%$ of Head and Neck (HN) cancer patients\cite{Strojan2017}. Treatments are planned on a patient’s computed tomography (CT) scan, where the tumour and the organs-at-risk are segmented. These segmentations are also used to establish dose-effect relationships which are ultimately used to improve radiotherapy practice. Modern techniques which allow the investigation of sub-volume dose effects rely on spatial normalisation to map the dose distributions between patients\cite{Palma2020}.
Examples for these associations in HN radiotherapy include radiation dose to the base of the brainstem and late dysphagia (problems swallowing)\cite{VasquezOsorio2023}, dose to the masseter muscle and trismus (limited jaw movement)\cite{Beasley2018}. In these examples, the authors used intensity-based non-rigid image registration (NRR) to indirectly establish the correspondence of the anatomy between different patients.
Improved spatial normalisation, using point-wise correspondences rather than NRR algorithms, would reduce uncertainties in outcome modelling applications. However, manually annotating pair-wise correspondences is a complex and time consuming task, rendering its practice unfeasible.

Another promising use of correspondences in radiotherapy applications is in the initialisation of structure-guided image registration methods.
Currently, spline-based registration relies on estimating correspondence based on distance criteria\cite{VasquezOsorio2009} and more advanced finite-element based models rely on a set of boundary conditions, e.g. based on structure curvature\cite{Cazoulat2016}. These structure-based registrations are particularly useful for cases with dramatic changes, such as registration of images before/after an intervention\cite{VasquezOsorio2014} or of images separated by a long time period (e.g. paediatric follow-up or re-irradiation settings). A model that can quickly and accurately identify corresponding points on sets of anatomical structures would be particularly effective for incorporation into other non-rigid image registration frameworks.


The aim of this study was to find a solution to automatically identify corresponding anatomical points on organs for radiotherapy applications. In this study, we took an established model for simultaneous correspondence and interpolation estimation in everyday 3D shapes, \textit{Neuromorph}\cite{Eisenberger2021}, and retrained it on biomedical data, specifically HN organ segmentations from planning CT scans. 
It has previously been shown that the performance of geometric learning models for tasks involving radiotherapy organ shapes can be dramatically improved by incorporating the associated CT scan imaging\cite{Henderson2022miccai}, an approach not attempted in previous correspondence literature\cite{Klatzow2022, Shi2021, Nie2022}. Therefore we extended \textit{Neuromorph} in two ways in an attempt to optimise its performance for this application: 1) by directly complementing geometrical features with learned image features, and 2) by adding a novel imaging loss function component.
The performance of our resultant correspondence models were compared to a NRR algorithm currently used for outcome modelling.

\section{Materials and method}

\subsection{Dataset}
An open-access dataset of 34 head and neck CT scans with segmentations of the brainstem, spinal cord, mandible, parotid and submandibular glands was used for this study\cite{nikolov2018}. The segmentations are highly consistent and followed international guidelines, having been produced by an expert and then audited by three observers and a specialist oncologist with at least four years of experience.

\subsection{Pre-processing}
\label{ssec:preprocessing}
The CT scans had a $\sim2.5\times1\times1$mm voxel spacing and were truncated at the apex of the lungs to ensure consistency in the length of the cervical section of the spinal cord. The marching cubes algorithm was used to generate 3D triangular meshes for each organ, which were then smoothed with ten iterations of Taubin smoothing. The meshes were simplified using quadric decimation to 3000 triangles for each of the organs apart from the submandibular glands which were simplified to 2000 triangles because of their smaller volume. The organ meshes were then optimised by iteratively splitting the longest and collapsing the shortest edges. The CT scans were rigidly aligned to a single reference patient using \textit{SimpleITK} 2.0.2. The computed transformations were applied directly to the mesh vertices to align the organ shapes thereby avoiding interpolation artefacts.

\subsection{Model}
The source model used in this study, \textit{Neuromorph}, was originally presented by Eisenberger et al.\cite{Eisenberger2021}. \textit{Neuromorph} is a geometric learning model which, when given two 3D triangular meshes, predicts corresponding points and a smooth interpolation between the two in a single forward pass. The model performs unsupervised learning which is crucial for our applications because of the scarcity of high quality 3D data labelled with point-to-point correspondences.

\begin{figure}[ht]
\includegraphics[width=\textwidth]{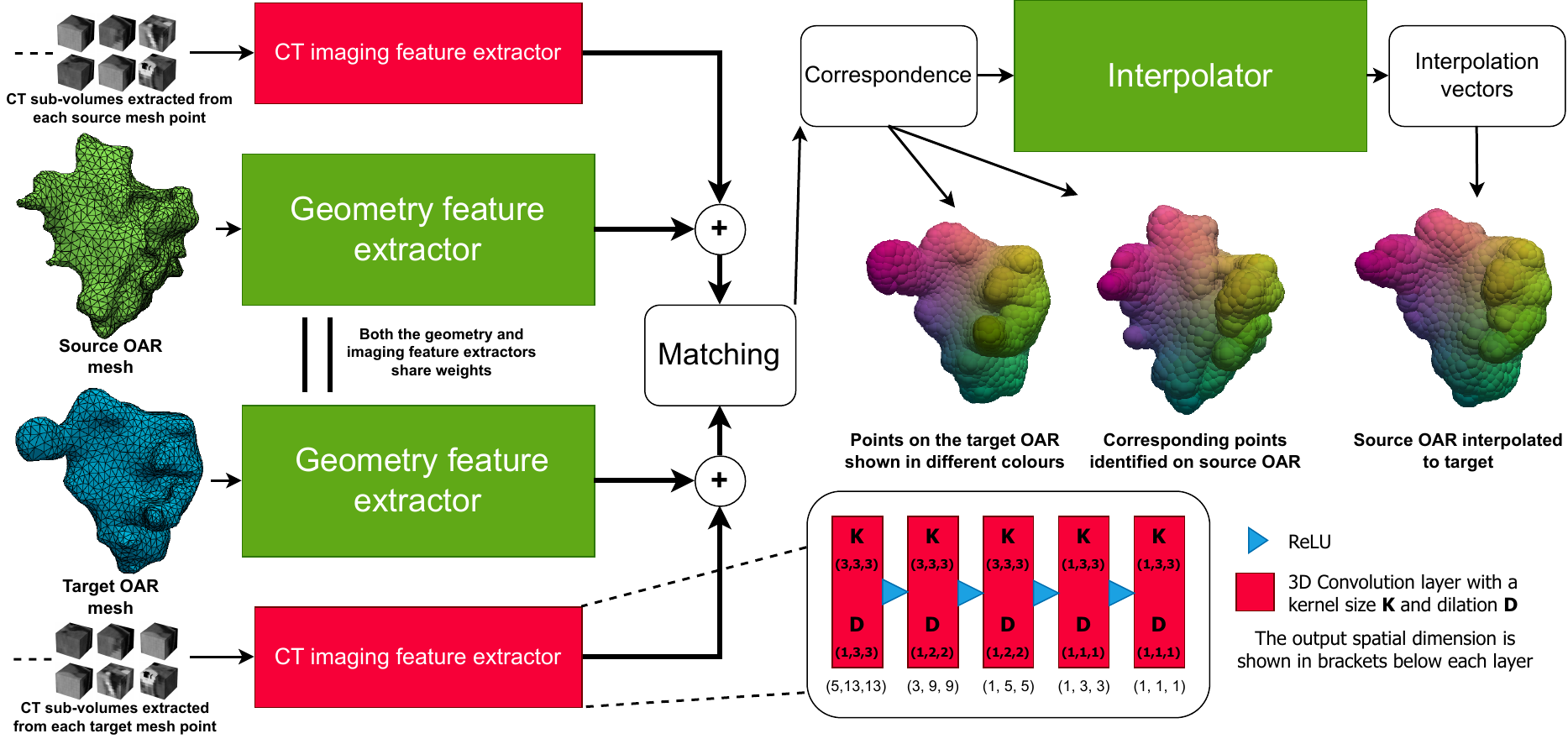}
\caption{A schematic of the \textit{Neuromorph} architecture\cite{Eisenberger2021} with our first extension to leverage the CT imaging. Additional CNN blocks (red) encode a $7\times19\times19$ CT sub-volume for each mesh point into an imaging feature vector to complement the geometrical features used to predict point correspondences. An example of the correspondence and interpolation predictions for a pair of parotid glands from different patients is shown with the different colours showing corresponding points.} \label{model_fig}
\end{figure}

Figure~\ref{model_fig} shows a schematic of the original model and one of our modifications to add imaging features. The \textit{Neuromorph} model is formed of two components: a Siamese feature extracting network and an interpolator. The feature-extracting portion consists of two networks with shared features which receive two meshes as input. The encoded shape features are matched using matrix multiplication to produce a correspondence matrix between the input meshes. The correspondence matrix is used to produce a vector which contains the offset between source vertices and their corresponding counterparts in the target mesh. This offset vector provides part of the input to the interpolator, along with the original source vertices and a time-step encoding to provide the number of intermediate steps along which to interpolate the deformation. The interpolator outputs a deformation vector for these time-steps for each vertex in the source mesh.

The feature extractor and interpolator have identical graph neural network architectures, consisting of repeating residual EdgeConv layers\cite{Wang2019}. The primary intuition behind success of the \textit{Neuromorph} architecture is that correspondence and interpolation are interdependent tasks that complement each other when optimised in an end-to-end fashion. \textit{Neuromorph} uses a three-component loss function for unsupervised learning. These are: a registration loss, to quantify the overlap of the target and source meshes; an ``as-rigid-as-possible'' (ARAP) loss, to  penalise overly elastic deformations; and a geodesic distance preservation loss, to regularise the predicted pair-wise correspondences. For all models in this study, the weight of the ARAP loss component was increased by a factor of ten compared to the originally proposed value to reduce the elasticity of predicted deformations. For full implementation details of the original \textit{Neuromorph} model, refer to \cite{Eisenberger2021}.

\subsection{Incorporating imaging information}
The original \textit{Neuromorph} model predicts point-to-point correspondences solely on the geometric structure of the input meshes. Since the meshes used in this study are organ shapes derived from CT scans, we have additional imaging information which was leveraged using two different approaches.

\subsubsection{Complementing geometrical features with image features} 
\begin{figure}[ht]
\centering
\includegraphics[width=\textwidth]{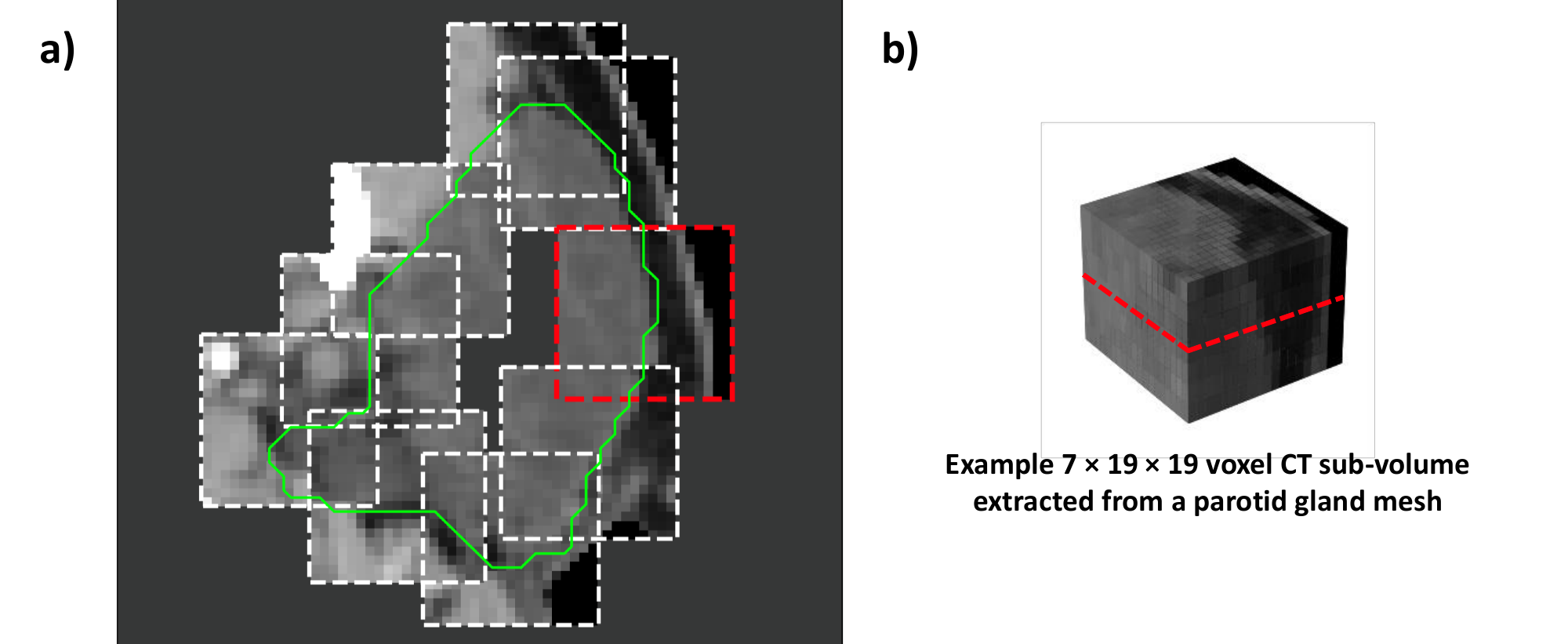}
\caption{a) A cross sectional view of a parotid gland mesh showing the field-of-view of the $7\times19\times19$ CT sub-volumes. Only $\sim10\%$ of the sub-volume patches in this cross section are shown for clarity of the visualisation. b) A visualisation of one of the 3D CT sub-volumes from the lateral aspect of the parotid.} \label{patches_fig}
\end{figure}

We followed a similar approach to that of Henderson et al. to encode image patches for each point on the mesh using a 3D convolutional neural network (CNN)\cite{Henderson2022miccai}. Figure~\ref{model_fig} shows further details of this methodology extension. Cubic 3D image patches of side length $\approx19$mm ($7\times19\times19$ voxel sub-volumes) were extracted from the CT scan for each vertex on the triangular mesh of each organ. This patch size was chosen so that image information 10mm outside the organ is within view, including surrounding structures such as bones and air cavities. Figure~\ref{patches_fig} shows an example slice of a parotid gland contour and demonstrates the field-of-view which these image patches cover. The image patches were normalised from Hounsfield Units (HU) onto the range $[0,1]$ using contrast windowing with settings used to visualise soft tissue (W 350HU, L 40HU)\cite{Hoang2010}.
The patches were then encoded using a custom CNN architecture into imaging feature vectors (Figure~\ref{model_fig}). These imaging feature vectors were concatenated with the feature vectors created by the geometric feature extractors of the original \textit{Neuromorph} model. Feature matching and correspondence prediction was then performed as before, but now utilising both geometric and imaging information. The imaging and geometric feature extractors of the extended model were optimised simultaneously during training. 

\subsubsection{Imaging as a component of the loss function}
\label{imaging loss}
For the second approach, we added a new loss term to calculate the mean-squared error of $7\times19\times19$ image patches for which the associated vertices are identified as corresponding. The $l_{\text{imaging}}$ loss component was calculated as

\begin{equation}
l_{\text{imaging}} = \lambda_{\text{imaging}} \times \|\Pi Y_{\text{CT\_patches}} - X_{\text{CT\_patches}}\|^2
\end{equation}

\noindent where $\Pi$ is the predicted correspondence matrix, $Y_{\text{CT\_patches}}$ and $X_{\text{CT\_patches}}$ are the CT image patches of the related target and source mesh points respectively and $\lambda_{\text{imaging}}$ was set to 1000 to balance the contribution with the other components. This hyperparameter value was chosen in preliminary testing from a range spanning $1 \rightarrow 100,000$.

By incorporating the imaging information as a loss component, the model does not require any additional input or modification from the original architecture. However, the rationale of including such an imaging loss was to encourage the model to learn more anatomically feasible correspondences at training time based on the underlying CT scan.

\subsection{Comparison with non-rigid image registration}
We compared the performance of the correspondence models with an established NRR algorithm which is a standard approach for aligning images and anatomical structures for radiotherapy applications\cite{VasquezOsorio2023, Beasley2018, Monti2017}. For this comparison, the CT scans were first rigidly registered to a single reference patient, as before, then \textit{NiftyReg} was used to non-rigidly register each pair of patients\cite{Modat2010}. The registration performed was a cubic B-spline using normalised mutual information loss with specific parameters: -ln 5 -lp 4 -be 0.001 -smooR 1 -smooF 1 -jl 0.0001. The computed non-rigid transformations were applied to the organ masks which were then meshed as in section~\ref{ssec:preprocessing}. Corresponding points between the pairwise registered organs were assigned using the nearest neighbours.

\subsection{Evaluation metrics}
\label{eval_metrics}
We implemented each of the three metrics used by Eisenberger et al.\cite{Eisenberger2021}:

\textbf{The geodesic error:} measures the consistency of shapes for sets of corresponding points\cite{Shilane2004}. It is defined as the differences between the geodesic distances of pairs of points on the target and the predicted corresponding pairs of points on the source mesh. This metric quantifies the discrepancies in the geodesic distances, resulting from the predicted correspondences, normalised by the square root area of the mesh. 

\textbf{The chamfer distance:} measures the accuracy of the predicted interpolation. It is defined as the distance between each predicted point on the source mesh to the nearest point on the target\cite{AkmalButt1998}. While the chamfer distance is a good measure of the overlap of the predicted shapes, a sufficiently elastic (and anatomically unrealistic) registration can achieve a near perfect (zero) chamfer distance.

\textbf{The conformal distortion:} provides insight into the realism of the deformations produced\cite{Hormann2000}. This metric quantifies the amount of distortion each triangle on the mesh experiences through interpolation. The conformal distortion is a good indicator of the anatomical feasibility of a deformation, with a higher conformal distortion metric value suggesting a more unrealistic registration.

\begin{figure}[ht]
\centering
\includegraphics[width=\textwidth]{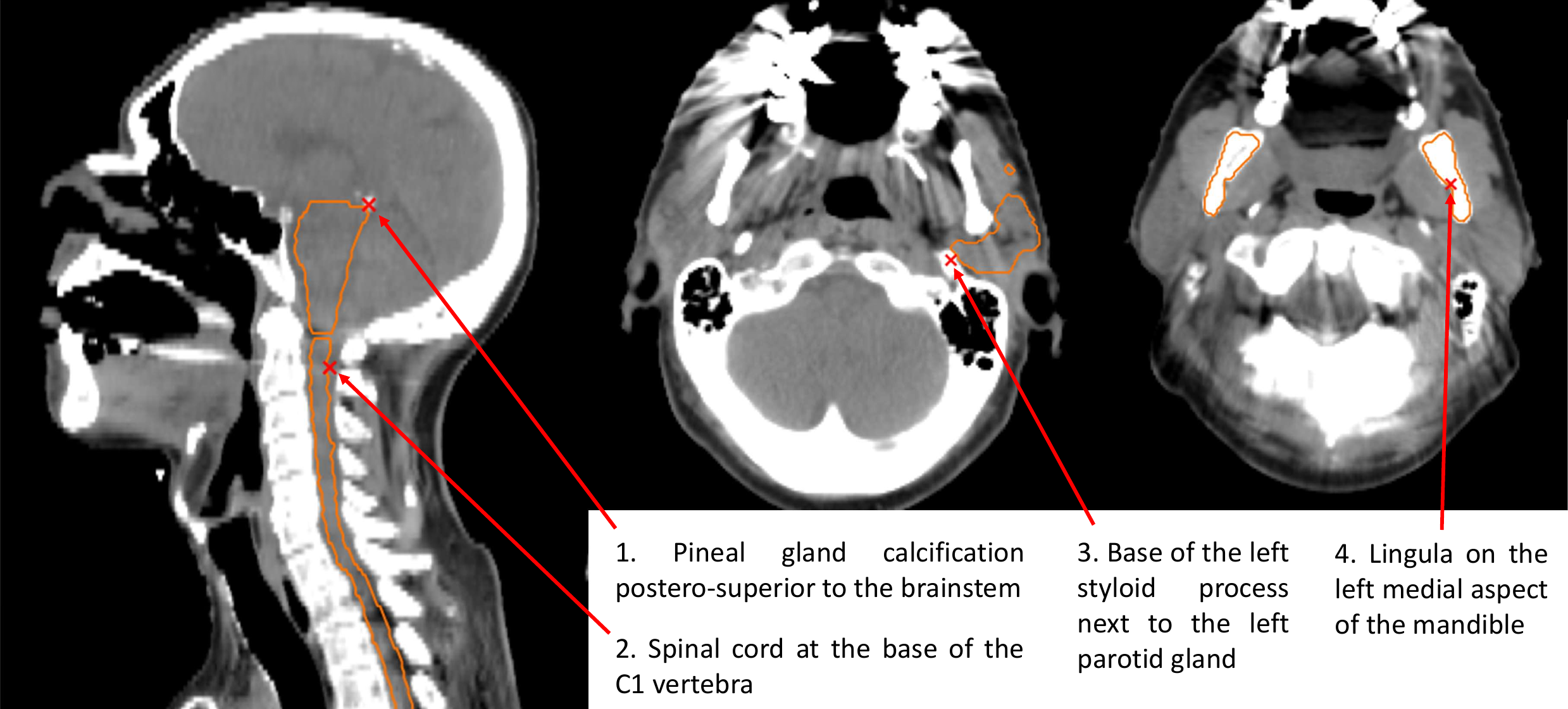}
\caption{CT scan slices showing the locations of the anatomical landmarks used for clinical validation.\vspace{-3mm}} \label{landmarks_fig}
\end{figure}

\subsubsection{Anatomical landmark error}
We additionally evaluated the correspondence of organ sub-regions using anatomical landmarks identified in the original CT scans. Figure~\ref{landmarks_fig} shows the landmarks used in this study which were manually identified in each of the 34 CT scans by a single observer. 
When the identified landmark was not on the segmentation, the closest point on each mesh was found. The Euclidean distance between the landmark on the target organ and the predicted corresponding landmark point was then found, which we call the landmark error.

\subsection{Implementation details}
All models are implemented using \textit{PyTorch} 1.13.0 and \textit{PyG} 2.2.0. \textit{Open3D} 0.13.0 and \textit{PyVista} 0.38.6 were used to perform mesh smoothing and visualisation. All training was performed using a 24 GB NVidia GeForce RTX 3090 and AMD Ryzen 9 3950X 16-Core Processor. The base \textit{Neuromorph} model contained $389,507$ parameters and the extended model with imaging features contained $686,467$ parameters.
Models were trained for 75 epochs with the Adam optimiser (learning rate of $0.0001$) and used a maximum of 4.8 GB GPU memory.

\subsection{Experiments}
For this study we evaluated the original model, \textit{Neuromorph}, and two proposed extensions against a NRR baseline. Each model was trained with data from all organs, but with the restriction that only pairs of the same organ were presented to the model, e.g., a pair of left parotid glands, followed by a pair of mandibles, etc. 
For each configuration we performed a five-fold cross-validation, dividing the data into folds to train five different model parameter sets. Trivial self-pairs were excluded when computing the evaluation metrics in section~\ref{eval_metrics}, resulting in $7\times24^2 = 4032$ pairs for training, $7\times3^2 = 63$ pairs for validation and $7\times7\times6 = 294$ pairs for testing each parameter set. The metric results in the testing fold for all five parameter sets are reported.

A Wilcoxon signed-rank hypothesis test was used to compare the performance of each of the model configurations to the NRR baseline for the anatomical landmark error. The geodesic error and chamfer distances were also calculated for the non-rigidly registered organs, but the conformal distortion could not be computed for the baseline approach since this metric requires a vertex-wise interpolation sequence.

\section{Results}
\begin{figure}[ht!]
\centering
\includegraphics[width=\textwidth]{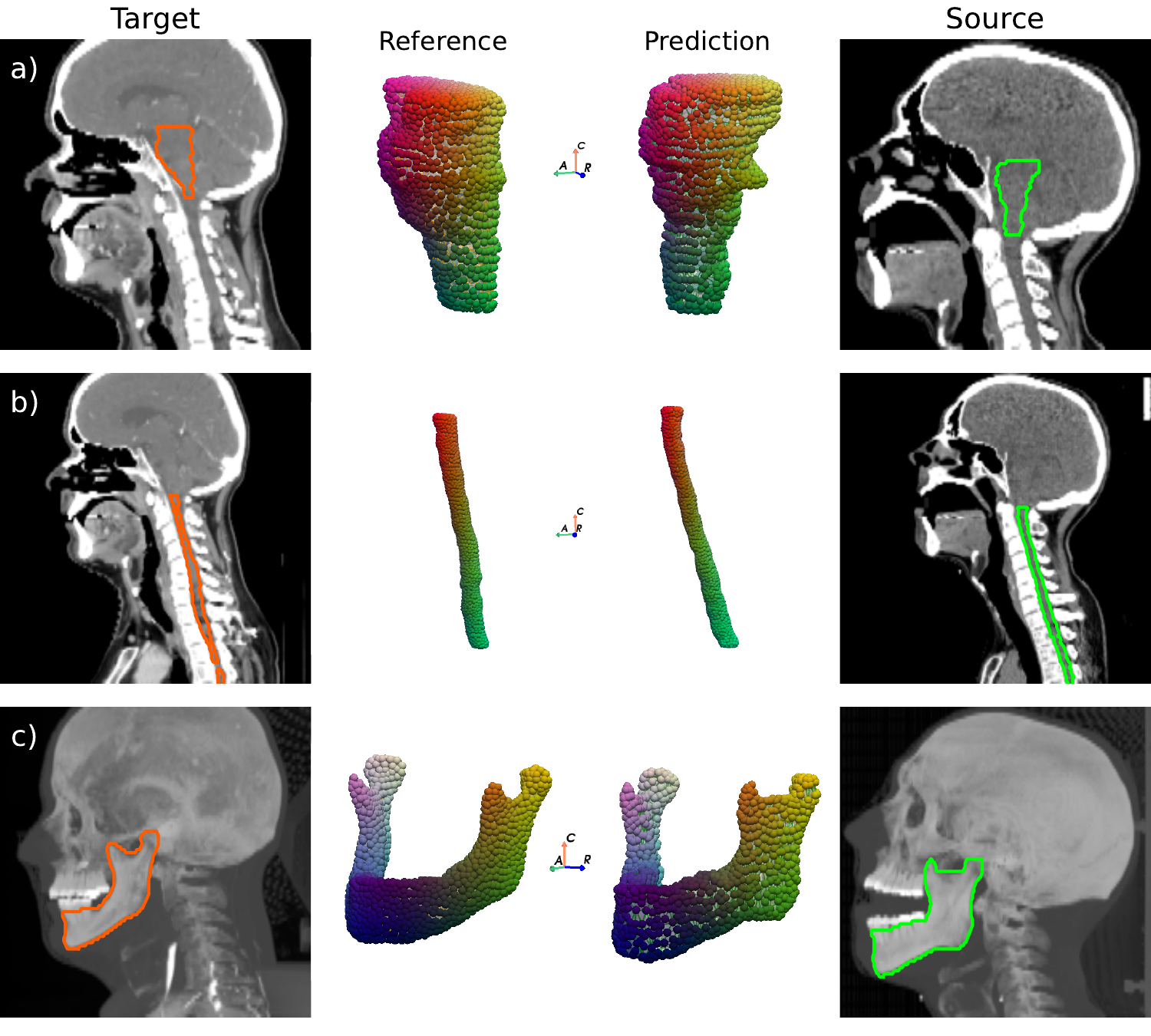}
\caption{Visualisation of predicted correspondences for a) the brainstem, b) the spinal cord and c) the mandible between a single pair of patients. The target patient scan and contour is presented in the first column, followed by the reference/target mesh, then the predicted correspondence on the source mesh, and finally, the scan and contour of the source patient. Sagittal slices or a maximum intensity projection (mandible) of the CT scans are shown to improve visualisation clarity.
\vspace{-5mm}} \label{results_visualisation_fig_a}
\end{figure}

\begin{figure}[ht!]
\centering
\includegraphics[width=\textwidth]{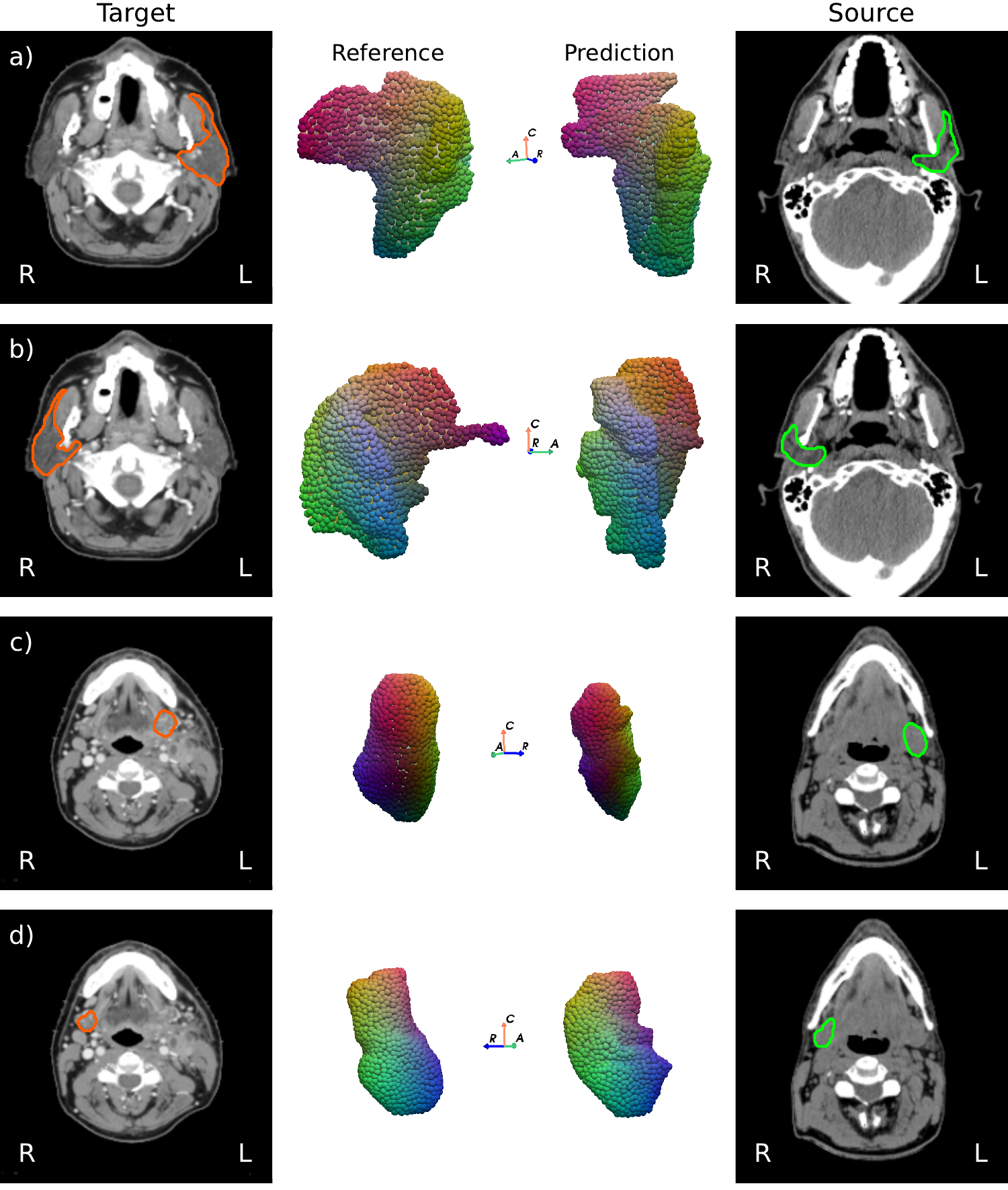}
\caption{Visualisation of predicted correspondences for a) the left parotid gland, b) the right parotid gland, c) the left submandibular gland and d) the right submandibular gland between a single pair of patients. The target patient scan and contour is presented in the first column, followed by the reference/target mesh, then the predicted correspondence on the source mesh, and finally, the scan and contour of the source patient. Axial slices of the CT scans are shown.
\vspace{-3mm}} \label{results_visualisation_fig_b}
\end{figure}

Figures~\ref{results_visualisation_fig_a} and \ref{results_visualisation_fig_b} show an example set of correspondence predictions for every organ between a single pair of patients. Identical colours on the organs identify corresponding points. 2D images, either axial or sagittal slices or maximum intensity projections of the CT scans are shown annotated with the organ contour to aid visualisation. The predictions shown were produced by a single model that included the imaging loss during training.

\begin{figure}[t]
\centering
\includegraphics[width=11.85cm]{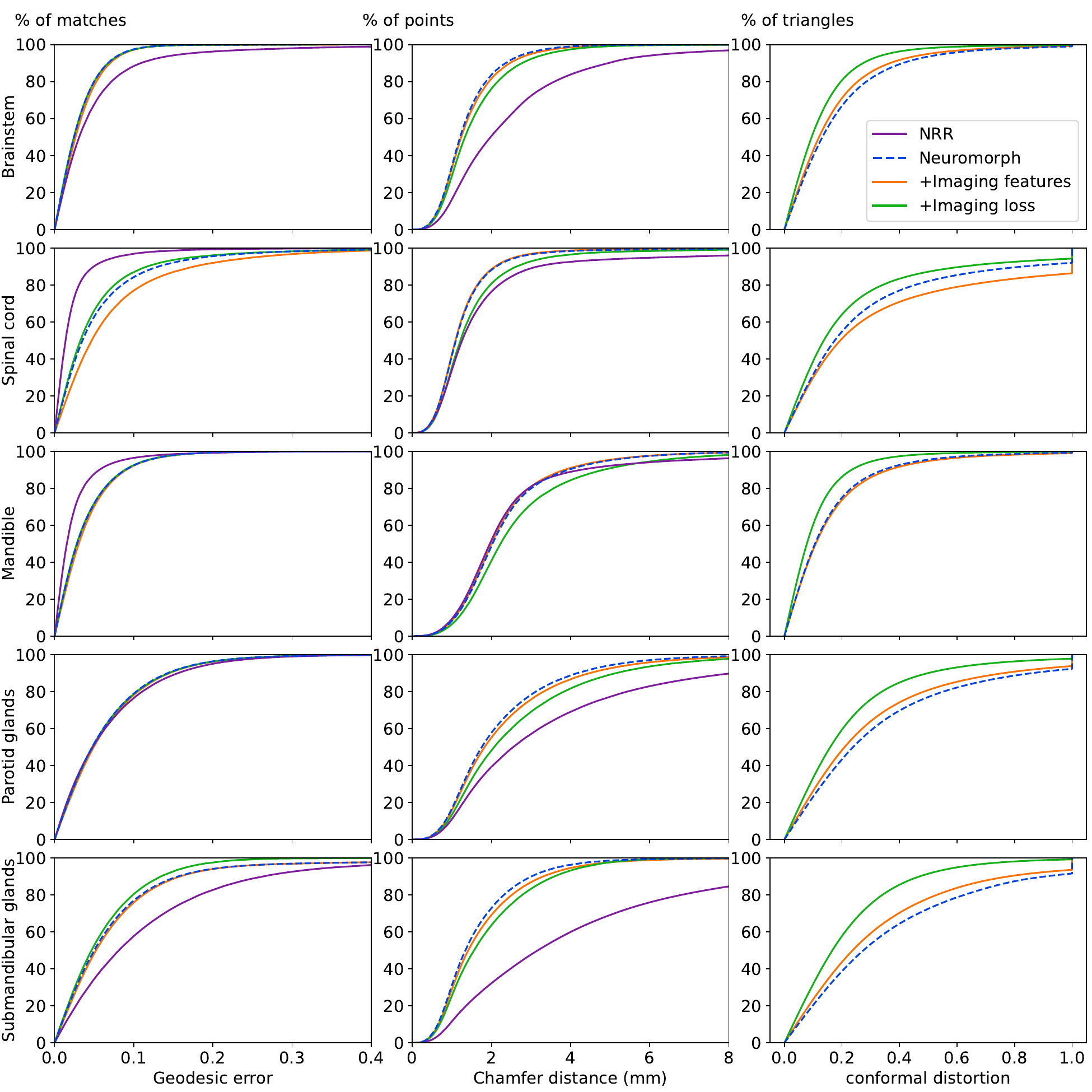}
\caption{Cumulative distributions of the geodesic error, chamfer distance and conformal distortion metrics for each of the model configurations and the NRR baseline. Closer to zero is better for all metrics.\vspace{-3mm}} \label{results_fig}
\end{figure}

Figure~\ref{results_fig} shows cumulative distributions of the geodesic error, chamfer distance and conformal distortion of all model configurations and organs. The original (\textit{Neuromorph}) model and model with imaging features perform similarly across most metrics. The original performs better on the geodesic error and conformal distortion for the spinal cord. However, the imaging features model produces less distortion for the parotid and submandibular glands.
The model which includes the imaging as an additional loss component performed similarly to the original for the geodesic error, apart from the submandibular glands for which it outperforms the original. The imaging loss model has a slightly poorer chamfer distance results compared to the original, but greatly improved conformal distortion results, especially for the submandibular glands.

For the geodesic error, the NRR baseline performs better than the correspondence models in the spinal cord and mandible, similarly for the parotid glands and worse for the brainstem and submandibular glands. The correspondence models outperform the NRR baseline for the chamfer distance for all organs apart from the mandible.

\begin{table}[ht!]
\caption{Landmark error distances for each model configuration. The level of significance of improvement of each model over the NRR baseline according to the Wilcoxon signed-rank test is shown as: * - p value $< 0.05$, $\textastdbltight$ - p value $< 0.005$, $\dag$ - p value $< 0.0005$, $\ddag$ - p value $< 0.00005$.}\label{landmarks_table}
\centering
\setlength\tabcolsep{1.2mm}
\begin{tabular}{lcccc}
& \multicolumn{4}{c}{Median landmark error (IQR) mm} \\
Model configuration & \multicolumn{1}{l}{Pineal gland} & \multicolumn{1}{l}{\begin{tabular}{@{}c@{}}Spinal cord \\ at C1\end{tabular}} & \multicolumn{1}{l}{Styloid process} & Mandible lingula \\ \hline
Baseline (NRR) & 4.6 (4.4) & 3.9 (3.6) & 8.4 (7.0) & 7.8 (10.8) \\[2mm]
Neuromorph & 3.7 (2.5) \dag & 2.3 (2.0) \ddag & 5.6 (5.3) * & 3.1 (2.1) \ddag \\
+ Imaging features & 3.7 (2.7) \dag & 2.2 (1.9) \ddag & 6.2 (5.8) * & 3.0 (2.4) \ddag \\
+ Imaging loss & 3.8 (2.5) \textastdbltight & 2.5 (2.2) \ddag & 6.3 (5.5) * & 3.1 (2.1) \ddag \\[2mm]
\begin{tabular}{@{}l@{}}Distance from \\ landmark to organ\end{tabular} & 3.6 (2.4) & 2.1 (1.6) & 5.4 (5.0) & 2.6 (1.8)
\end{tabular}
\end{table}

Table~\ref{landmarks_table} shows the landmark error distances for each of the model configurations and the NRR baseline. All of the models showed a significant improvement over the baseline for all anatomical landmarks. All correspondence methods perform similarly in terms of landmark distance, but subtle differences could exist that are hidden by observer variation. The median distance from the landmark to the organs is shown in the final row and this serves as an indication on the landmark variability and hence a reasonable upper bound of the correspondence accuracy identifiable with this measure.

\begin{figure}[ht!]
\centering
\includegraphics[width=\textwidth]{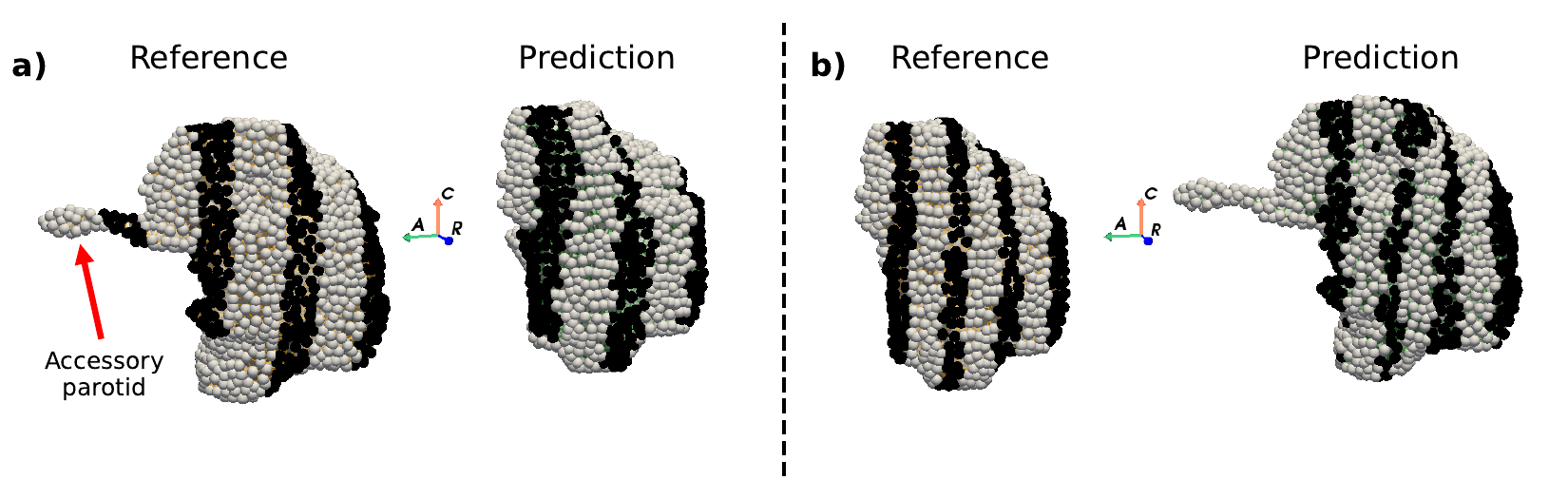}
\caption{An example of the model including imaging loss robustly handling a case with missing correspondences for a pair of parotid glands. In a) the reference parotid has an anterior extension (accessory) which is not reproduced on the predicted correspondences. In b) the lack of accessory in the reference does not impact majority of the predicted correspondences shown by the black stripes aligning between the two. Black and white has been used here to show corresponding points instead of the full colormap for clarity.}\label{robustness_fig}
\end{figure}

Figure~\ref{robustness_fig} shows an additional example of correspondences produced by the model including imaging loss. This particular case is interesting as it demonstrates how the model handles the difficult scenario of missing correspondences. One of the patients has an accessory parotid, an anterior extension of the parotid present in \textgreater30\% of the population \cite{Rosa2020}, and the other does not. The model was able to robustly handle this case in both directions, i.e. with either patient as the reference.

\section{Discussion}
In this study we showed that an established neural network for predicting correspondence and smooth interpolation of 3D shapes can be applied to HN organ segmentations from CT scans. We additionally evaluated two methodological extensions to leverage the CT imaging information.

The correspondence models were compared to an intensity-based NRR algorithm regularly used for radiotherapy outcome modelling. The NRR produced better correspondences for the spinal cord and mandible in terms of the geodesic error showing the effectiveness of the image registration method for the more straightforward task of aligning the skeleton and anatomy enclosed by bone. However, the original \textit{Neuromorph} model and extensions all produced significantly lower landmark errors for every organ than the NRR baseline as well as producing better chamfer distance results for soft tissue organs. 
This promising result demonstrates the potential of such correspondence methods to reduce uncertainties in radiotherapy outcome modelling. Further work is required to quantify the uncertainty reduction and its impact for this purpose.

An intensity-based non-rigid registration algorithm was used as a comparison baseline for the learning-based correspondence models. A mesh-based registration such as coherent point drift could have alternatively been applied for a more direct comparison\cite{Myronenko2010}. However, intensity-based image registration algorithms are the current standard for aligning images and structures for radiotherapy applications, particularly for outcome modelling, and therefore provide a more relevant comparison for this study\cite{VasquezOsorio2023, Beasley2018, Monti2017}.

The performance of the original \textit{Neuromorph} model was slightly improved by incorporating imaging information, not as explicit imaging features, but rather by introducing an additional imaging term to the loss function. This configuration does not require imaging at inference time, instead the imaging is used solely when training. This configuration was shown to be particularly effective at regularising the predicted correspondences with substantially reduced conformal distortion results. This indicates that the inclusion of an imaging loss term produced more anatomically feasible and robust deformations. The improvement in conformal distortion metric, whilst hardly affecting performance in terms of the other metrics, makes this particular configuration appealing for future exploration as a starting point for an anatomically informed non-rigid registration method.

The original \textit{Neuromorph} model was also extended to receive imaging input directly and predict correspondences based on geometric and imaging features, but this extension did not improve performance. We believe that this is primarily due to the highly consistent data used for model training. The segmentations were as close to the consensus guidelines as possible which is unlikely in clinical practice. This meant the contours will deviate only slightly from ``true'' organ boundaries. We envisage providing the imaging information as input to the model to be of greater use in scenarios where the segmentations are more variable, and could be inconsistent with the underlying anatomy. This is an interesting avenue for future work.

While \textit{Neuromorph} is an established model for everyday 3D shapes, we believe this is the first time it has been shown to be effective in biomedical applications. Additionally, while there are other learning based correspondence methods\cite{Klatzow2022, Attaiki2023}, this is the first to combine geometric learning and leverage imaging, providing a slight improvement on the original model in terms of anatomical feasibility.


The additional imaging loss component described in section~\ref{imaging loss} utilises the mean squared error for simplicity. This metric is only appropriate when quantifying the similarity of mono-modal scans which are intensity calibrated, such as CT scans used for radiotherapy planning. If the underlying imaging was a cone-beam CT or MRI, an alternative measure such as mutual information or correlation ratio could be used.

Our model was primarily developed with outcome modelling in mind, which relies on inter-patient analysis. Inter-patient correspondence is a more complex task than identifying intra-patient correspondence since there is greater variability in the anatomy. Consequently, we believe that extension to the intra-patient tasks should be straightforward.

\section{Conclusion}
We have shown that an established model, originally developed for generic 3D shapes can be adapted for applications in biomedical imaging. Specifically, this model could be used to identify corresponding points on 3D organs to improve spatial normalisation in outcome modelling applications, potentially reducing the associated uncertainties and facilitating the development of better radiotherapy treatments. Further, we envision that in the future, such a correspondence tool, which also provides a smooth interpolation, could be deployed at the heart of an effective, anatomically informed non-rigid registration method.

\newpage

%
%
%
\bibliographystyle{splncs04}
\bibliography{bibliography}

\end{document}